\documentclass{article}

\usepackage{geometry}
\usepackage{graphicx} 
\usepackage{makecell}
\newcolumntype{C}{>{\centering\arraybackslash}X}  
\usepackage{amsmath}
\usepackage{amssymb}
\usepackage{booktabs}
\usepackage{tabularx}
\usepackage{adjustbox}
\usepackage{subcaption}
\usepackage{array}
\usepackage{tikz}
\usepackage{nicematrix}
\usepackage{url}
\usepackage{multirow}
\usepackage{caption}

\usepackage{float}
\usepackage{booktabs}
\usepackage{authblk}
\usepackage{xcolor}
\definecolor{Class_0!}{RGB}{0, 0, 0} 
\definecolor{Class_1!}{RGB}{255, 0, 0} 
\definecolor{Class_2!}{RGB}{70, 130, 255} 
\definecolor{Class_3!}{RGB}{255, 255, 0} 
\definecolor{Class_4!}{RGB}{255, 0, 255} 

\usepackage{color}
\definecolor{blue!}{RGB}{79, 121, 167}
\definecolor{red!}{RGB}{225, 87, 90}
\definecolor{green!}{RGB}{118, 183, 179}
\newcommand{\colorbar}[1]{%
  \textcolor{#1}{\rule[0ex]{1.1ex}{1.1ex}}%
}

\title{MWIR-4-Plastic: The Identification of Complex End-of-Life Industrial Plastic using Mid-wave Infrared Hyperspectral Imaging and  Machine Learning}
\author{\textsuperscript{1,2}Elias~Arbash, 
\textsuperscript{1}Andréa~de~lima Ribeiro, 
\textsuperscript{1}Filipa~Simões, 
\textsuperscript{1}Ahmed~Jamal~Afifi, 
\textsuperscript{1,3}Aldino~Rizaldy,
\textsuperscript{1}Yuleika~Madriz, 
\textsuperscript{1}Samuel~Thiele, 
\textsuperscript{1}Sandra~Lorenz, 
\textsuperscript{1}Margret~Fuchs,
\textsuperscript{1,4}Pedram~Ghamisi,
\textsuperscript{2}Paul~Scheunders,
\textsuperscript{1,5}Richard~Gloaguen\\
\textsuperscript{1}Helmholtz-Zentrum Dresden-Rossendorf (HZDR) -\\ Helmholtz Institute Freiberg for Resource Technology (HIF), Freiberg, Germany\\
\textsuperscript{2}University of Antwerp, Antwerp, Belgium \\
\textsuperscript{3}Remote Sensing and Geoinformatics, Freie Universität Berlin, Berlin, Germany\\
\textsuperscript{4}Faculty of Electrical and Computer Engineering, University of Iceland, 101 Reykjavik, Iceland\\
\textsuperscript{5}TU Bergakademie Freiberg, Germany\\
}
\date{}

\begin{document}

\maketitle

\begin{abstract}

The automated sorting of shredded black plastics from end-of-life (EOF) industrial waste presents a significant challenge in recycling facilities, primarily due to the limitations of current sensing and analytical approaches. Existing studies predominantly rely on single-point contact-based mid-infrared spectroscopy or laboratory hyperspectral imaging (HSI) setups, which fail to provide the spatially resolved analysis necessary for fast, bulk processing. 
Moreover, available datasets are laboratory-controlled and focus on intact rather than shredded plastics, hindering further recycling refinement. Black industrial plastics, in particular, are underrepresented, while most classification pipelines depend on manual region selection and rule-based spectral matching, neglecting spatial information and modern deep learning (DL) methods.
To address these gaps, we introduce the first publicly available HSI dataset of shredded black plastics from EOF vehicle, comprising four industrial polymers across 13 co-registered RGB, VNIR, SWIR, and MWIR scenes and their segmentation pipeline.
We developed a multi-modal spectral-spatial framework that integrates foreground isolation, pixel-wise classification, and object-level majority voting. By adapting advanced hyperspectral transformers from earth observation and incorporating chemometric band selection, we achieve accurate classification of complex black plastics. 
The study establishes the first comprehensive benchmark using nine processing methods, including chemometric, machine learning, and DL architectures. To ensure reproducibility, the complete dataset and methodologies are publicly released, establishing a benchmark for a hyperspectral object-analysis pipeline in industrial inspection.

\end{abstract}

\section{Introduction}
\label{sec:intro}

Global plastic pollution has escalated into an urgent environmental crisis. Annual plastic production stands at approximately 430 million tons per year and is projected to exceed 30 billion tons by 2050 \cite{owid-plastic-pollution, li2022expanding, yang2025application}. Much of this material leaks into marine and terrestrial ecosystems, forming plastic pollutants that weaken soil fertility, damage habitats, and threaten biodiversity \cite{kalali2023critical, stoett2024global}. Traditional waste disposal methods often contribute significantly to environmental degradation, including microplastic accumulation in oceans and freshwater systems and greenhouse gas emissions from incineration  \cite{liu2023automated}. 
Effective recycling strategies are thus needed to mitigate these ecological harms and promote sustainable resource use and management. \cite{meijer2021more, de2024multi}. 

While the packaging sector dominates plastic consumption, accounting for roughly 39\% of the European market, plastic waste from the building and construction as well as automotive industry accounts for $\sim 21\%$ and $\sim 9\%$, respectively, in addition to $\sim 7\%$ contributions from waste from electrical and electronic equipment (E-waste or WEEE) \cite{europe2024circular, sormunen2022towards}. Unlike packaging plastics, which are subject to strict food-safety regulations and contain fewer hazardous additives, plastics from the industrial and post-consumer sectors commonly contain additives that change mechanical properties, colour, durability, and/or flame retardancy \cite{kato2021safety}. These additives severely complicate end-of-life processing, typically resulting in downcycling and degradation of reprocessed polymers \cite{eriksen2018contamination}. 
Poorly sorted recycling streams drastically increase reprocessing costs and decrease the market value of the resulting products; therefore, precise separation of plastics according to their composition is a necessary step to increase recycling rates, as reported by \cite{de2024multi, sormunen2022towards, stenmarck2017hazardous}.

Conventional plastic sorting technologies such as gravimetric separation, air classification, electrostatic separation, and manual sorting suffer from several limitations \cite{yang2025application, carrera2022machine}: they all rely on differences in the mechanical or physical properties of materials, or require chemical reagents. However, in complex plastic streams, different polymer compositions often exhibit similar densities, undermining density-based separation techniques such as flotation. This results in secondary contamination, increased energy consumption, health risks for operators, and limited scalability for highly heterogeneous waste streams \cite{singh2024plastic, lubongo2022assessment}. 
Refined sorting preserves the integrity of recycled materials, avoiding cross-contamination of different polymer grades \cite{hahladakis2018overview}.

Imaging-based systems enable fast characterization of moving streams without physical contact \cite{de2024multi}. Among these technologies are true color (RGB) and HSI cameras, which provide spatial and compositional information within a single measurement, enabling near real-time mapping of material stream compositions \cite{Roming2026MWIRBlackPlastics}. 
Current industrial hyperspectral sensors primarily operate in the visible near-infrared (VNIR, 400--1000 nm) and short-wave infrared (SWIR, 1000--2500 nm) spectral regions \cite{bonifazi2023hyperspectral, serranti2024efficient, taneepanichskul2024using}. Polymer identification in these wavelength ranges, particularly within 1600--2200 nm, is based on characteristic absorption features originating from molecular vibrational overtones and combination bands, commonly referred to as spectral fingerprints and organic proxies \cite{de2024multi}. 

Accordingly, several studies showed that SWIR-HSI data allow the detection of specific light-coloured polymer fingerprints, enabling the identification of major industry-relevant polymer types. Bonifazi et al. \cite{bonifazi2023hyperspectral} used an Imspector N25E Specim camera (1000--2500 nm) with chemometric methods for identifying eight polymer types of light-coloured fractions from EOF electronic waste: polyoxymethylene (POM), polymethyl methacrylate acrylic (PMMA), polycarbonate (PC), polystyrene (PS), acrylonitrile butadiene styrene (ABS), polypropylene (PP), polyethylene (PE) and polyamide (PA). 
Further study by Bonifazi et al. \cite{bonifazi2021detection} used an ASD FieldSpec® 4 Standard-Res field portable spectroradiometer (350--2500 nm), and an Imspector N25E Specim camera in the SWIR range to detect bromine content and discriminate ABS from PS — two polymer types of very similar spectral features — in shredded plastic scraps from cathode ray tube monitors and televisions.
Serranti et al.\cite{serranti2024efficient} also used an Imspector N25E Specim camera with different lenses to evaluate the influence of spatial resolution, spectral range (1000--1700 nm versus 1000--2500 nm), and classification models for the characterization of micron-scale microplastics, demonstrating the feasibility of identifying polymers such as PS, PP, and high-density polyethylene (HDPE). 
Similarly, Taneepanichskul et al. \cite{taneepanichskul2024using} used a HySpex BaldurS-640iN hyperspectral camera (950--1730 nm) with chemometric methods to distinguish compostable plastics, including polylactic acid (PLA) and polybutylene adipate terephthalate (PBAT), from non-compostable polymers such as PP, polyethylene terephthalate (PET), and low-density polyethylene (LDPE) in industrial composting streams.


Despite their widespread adoption, VNIR and SWIR sensors are primarily limited to the detection of non-dark plastics. The presence of carbon black and other infrared-absorbing pigments significantly decreases the intensity of signals reflected by the polymer, reducing the signal-to-noise ratio and masking the overtone absorption features used for polymer discrimination \cite{vazquez2015multi, xia2021convolutional}. VNIR-SWIR sensors are thus not suitable for the classification of dark and black plastics, which constitute a substantial fraction of WEEE and EOL vehicle streams \cite{Roming2026MWIRBlackPlastics, yang2025application}. 
A promising alternative is to exploit the Mid-Wave Infrared (MWIR, 3000--5000 nm) spectral region, where molecular spectra are dominated by larger-amplitude fundamental vibrational modes rather than their weaker overtones, yielding distinctive, chemically discriminating absorption patterns even for visually similar black polymers \cite{rozenstein2017development, vazquez2015multi}.
Moreover, MWIR measurements are considerably less affected by carbon black pigments, enabling reliable spectral acquisition from dark and black plastics that are difficult to characterize using VNIR or SWIR sensors \cite{de2024multi, signoret2020mir}. When coupled with chemometric and ML methods, HSI-MWIR thus becomes a promising sensing modality for next-generation automated recycling systems  \cite{Roming2026MWIRBlackPlastics}.

Early studies established that spectral ranges beyond the SWIR are more informative for polymer identification, offering more robust discriminative features than VNIR or SWIR regions alone; this was primarily demonstrated using in-contact Fourier-transform infrared (FTIR) point spectroscopy operating in the mid-infrared spanning 2500--14000 nm depending on instrument configuration.
Using an FTIR spectrometer in that range, several polymer investigation studies were conducted: Vazquez et al. \cite{vazquez2015multi} used an FTIR spectrometer to develop a spectral fingerprint library for twelve plastic resins including PET, HDPE, PVC, LDPE, PP, PS, and PC from household and end consumer products, demonstrating that carbon-black pigmentation does not influence MWIR spectral signatures. Another study is Rozenstein et al. \cite{rozenstein2017development} who further confirmed using FTIR spectroscopy the capability of the 2500--14000 nm range over the VNIR range in identifying coloured, transparent and black plastic waste including PET, PE, PVC, PP, PLA and PS, in addition to the ability of sorting general post-consumer waste such as wood, paper, and plastics. Yang et al. \cite{yang2025application} also used an FTIR spectrometer with nine chemometric methods for the classification of 698 samples from 9 polymer types, including various sizes and colours of waste objects such as black, white and transparent PE, PA, PS, PC, ABS, PET, polybutylene terephthalate (PBT), PP and PVC.
In the third installment of a series characterizing spectra of WEEE plastics for industrial sorting, Signoret et al. \cite{signoret2020mir} investigated with laboratory FTIR-ATR spectroscopy in the range 2500--25000 nm how accelerated aging affects styrenics more than polyolefins and impacts long-wave infrared (LWIR) sub-ranges more than MWIR.
Using a Vertex 70 FTIR spectrometer, Stavinski et al.\cite{stavinski2023mid}  classified PET, HDPE, LDPE, PP, and PS post-consumer plastic waste within the range 2500--15400 nm, applying autoencoders together with ML classifiers, and identified the $C–H$ stretching region as free of the additive- and dye-related interference that limits conventional VNIR sorting.
Following these feasibility studies beyond the SWIR range, further progress was made using MWIR-HSI: De Lima et al. \cite{de2024multi} conducted multi-sensor investigations using both handheld sensors, including FTIR, Raman, and HSI to compare the abilities of VNIR, SWIR, MWIR, and LWIR in identifying nine E-waste plastic types including HDPE, LDPE, PP, PET, PVC, PC, PS, ABS and PMMA, demonstrating the complementary information provided by MWIR for black polymer identification while revealing limitations of individual sensing technologies. More recently, Roming et al. \cite{Roming2026MWIRBlackPlastics} demonstrated, using a Specim FX50 HSI camera and chemometric and machine-learning methods, the ability of MWIR-HSI to identify multiple polymer classes, such as HDPE, LDPE, PET, PP, and PS, from non-shredded domestic packaging waste. The study highlighted the spectral differences between coloured and carbon-black plastics of identical polymer composition. 





Across the reviewed literature, MWIR approaches have consistently been identified as a promising route to overcome the black-plastic sorting limitations of near-infrared-HSI \cite{signoret2020mir}.  In practice, this potential has been adopted by closed industrial sorting systems (e.g., the STEINERT Unisort BlackEye) \cite{REICHERT2026115408}. Nonetheless, their classification logic, spectral processing, and performance are proprietary and not disclosed or systematically benchmarked in the open scientific literature. This limits their value as a basis for reproducible scientific comparison and leaves MWIR-based classification largely underdeveloped in peer-reviewed, reproducible research, with existing studies constrained by several bounds:

\begin{itemize}
    \item Limited spatial automation: most studies rely on single-point contact-based MWIR spectroscopy or laboratory setups. Moreover, where imaging is used, classification pipelines still depend on manually selected regions of interest or hand-drawn masks rather than automatic object localization and segmentation, preventing the spatially resolved, fast, bulk analysis required for industrial sorting.

    \item Lack of representative industrial datasets: existing datasets are typically small, collected under controlled laboratory conditions, and predominantly consist of intact plastic objects instead of shredded post-consumer streams encountered in recycling facilities.

    \item Generic polymer identification: previous investigations mainly consider generic consumer polymers, while black engineering plastics originating from end-of-life vehicle (ELV) waste remain largely unexplored.

    \item Conventional data analysis methods: black polymer identification is predominantly based on rule-based spectral matching, handcrafted spectral descriptors, chemometric approaches (e.g., PCA, PLS-DA), or conventional ML models, whereas modern DL architectures remain largely unexplored for MWIR hyperspectral plastic sorting.

    \item Limited reproducibility: publicly available datasets, preprocessing pipelines, trained models, and inference frameworks are generally unavailable, hindering reproducible research and fair benchmarking.
\end{itemize}

We build on the earlier investigations by adopting state-of-the-art (SOTA) sensing (SWIR- and MWIR-HSI) and data processing technology for the identification of shredded complex polymers by:

\begin{itemize}
    \item Introducing the first publicly available HSI dataset of shredded black plastics from ELV, comprising 13 co-registered RGB, VNIR, SWIR, and MWIR scenes for four engineering polymer types with pixel-wise annotations, and semantic and instance segmentation masks.
    
\item Developing automated, multi-modal spectral-spatial processing pipelines that integrate complementary imaging modalities and combine foreground isolation, pixel-wise classification, and object-level majority voting for robust, automatic, and reliable identification of shredded plastic fragments.

    \item Adapting advanced hyperspectral transformers from earth observation to industrial recycling by incorporating chemometric band selection and data engineering techniques into the workflow for accurate classification of complex black plastics.

    \item Establishing the first comprehensive benchmark on industrial shredded black plastics, evaluating nine methods spanning rule-based spectral-matching baseline (SAM), classical multivariate and ML classifiers (LDA, SVM, Random Forest, KNN), and SOTA DL architectures (1D-CNN, ViT, SpectralFormer) for hyperspectral classification. Moreover, we release the complete preprocessing, training, and inference workflow as a fully reproducible pipeline for automated black-plastic identification and generalizable hyperspectral object analysis in other industrial inspection tasks.

\end{itemize}

This paper is organized as follows: Section 1 introduces the topic and previous work in plastic classification using spectroscopy-based sensors. Section 2 narrates the experimental work on the dataset, covering its description and every data processing step starting from acquisition, to methodologies and post-processing. Section 3 provides pixel-, object-, and instance-level evaluations, performance analysis, and limitations highlighting. Section 4 summarizes key findings and contributions.

\section{Experiments}
\subsection{Dataset: Samples and Acquisition }
We have acquired 13 hyperspectral scenes of post-shredding and post-sorting ELV plastic particles. These particles underwent conventional industrial processing, including the removal of metals and other non-polymeric components, followed by successive sorting stages using the STEINERT Unisort BlackEye machine for the separation of target polymer classes, as described in detail in \cite{REICHERT2026115408}. The sample arrangements were thus designed to capture specific stages of the current industrial sorting workflow. 
Two scenes contain mixed polymer particles, in which several distinct polymer classes remain present. They represent an ELV sorting product enriched in polymer particles before type separation. The remaining eleven scenes represent each polymer concentrate obtained after sensor-based sorting, including PP, PC, PA, and ABS-based materials (hereafter referred to as 'Styrene'). The complete dataset comprises more than 15,000 shredded automotive particles, most of which are black. Particles were distributed across two nominal size fractions with object diameter ranging approximately in the first one from 5 mm to 11 mm and from 5 mm to 30 mm, with particle counts ranging from 500 (single-class scenes) to 2000 (mixed-class scenes) particles. 
Data acquisition was performed at the Helmholtz Institute Freiberg for Resource Technology using a specialized SisuRock hyperspectral scanner (Specim, Oulu, Finland) operating at a fixed working distance. The platform contains three line-scan sensors (detailed in Table \ref{tab:parameters}): a JAI LT-400 CL RGB camera acquiring images at 160 micron sampling distance (ca. 6-times the spatial resolution of the hyperspectral sensors), an AisaFENIX camera covering the VNIR and SWIR spectral regions, and a Specim FX50 camera covering the MWIR region. The images were subsequently coregistered and downsampled to achieve a common spatial resolution of 1 mm/pixel across all modalities. 



\begin{table*}[t]
\caption{The parameters of the acquisition sensors and the selected processing bands along with their physical phenomena.}
\label{tab:parameters}
\centering
\small
\setlength{\tabcolsep}{4pt}
\begin{tabular}{|p{3.2cm} |p{3.2cm} |p{3.2cm} |p{3.2cm} |p{3.2cm}|}
\hline
\textbf{Parameter / Property} & \textbf{Specim AsiaFENIX (VNIR)} & \textbf{Specim AsiaFENIX (SWIR)} & \textbf{Specim FX50 (MWIR)} & \textbf{JAI LT-400 CL}\\
\hline
\textbf{Wavelength range (nm)} & $400–970$& $970–2500$ & $2700–5300$ & RGB \\
\hline
\textbf{Bands} & 175 & 275 & 308 & 3 \\
\hline
\textbf{Spectral resolution (nm)} & 3.5 & 12 & 35 & -- \\
\hline
\textbf{Spatial sampling (mm)} & \multicolumn{2}{|c|}{1.6}& 1.1 & 0.16 \\
\hline
\textbf{Illumination source} & \multicolumn{2}{|c|}{Halogen light bulb}& Shielded Ni/Cr resistance wire & -- \\
\hline
\textbf{Calibration panel} & \multicolumn{2}{|c|}{Zenith Polymer Diffuse $>99\%$}& Coarse Aluminium, $\sim 93\%$ & -- \\
\hline
\textbf{Integration times (ms)} & 18 & 4.5 & 0.75 & -- \\
\hline
\textbf{Regions of interest (nm)} & -- & $[1600–2200]$ & $[2976–3018]$, $[3336–3420]$, $[3428–3512]$, $[3663–3696]$, $[4458–4475]$ & -- \\
\hline
\textbf{Processed bands (index)} & -- & $[285–392]$ & $[32–38]$, $[75–85]$, $[86–97]$, $[114–119]$, $[209–212]$ & -- \\
\hline
\textbf{Range assignment} & -- & 1st overtone of CH stretching -- 2nd overtones of C=O/OH -- Combination bands of NH, OH, CH & Fundamental NH / OH stretching -- Aromatic CH stretching -- Aldehydic/Aliphatic CH -- $C\equiv N$ stretch in nitrile from SAN & -- \\
\hline
\end{tabular}
\end{table*}


\subsubsection{Ground Truth Assignment}
Although the industrial sorting process yields highly enriched class-specific polymer streams, these fractions are not entirely free of misclassified particles (e.g., non-polymeric materials or polymer materials from other classes) and can inject noise in the ML models' training data. 
Therefore, a pixel-level ground truth was established for each class based on the detection of diagnostic features reported in the literature. The analytical spectral ranges were selected according to the characteristic molecular vibration of polymers within the spectral range (Table \ref{tab:parameters}). 
The workflow for fingerprint detection based on minimum wavelength mapping (MWL) analysis is described in detail in \cite{de2024multi}. For each selected spectral range, MWL feature extraction was carried out on a pixel-wise basis, followed by decision tree classification for each hyperspectral scene using the Python-based Hylite toolbox \cite{thiele2021multi}. While this yields accurate results, its computational expense and sensitivity to manually selected thresholds hinder its application to industrial sorting. 

\subsection{Preprocessing}

Data preprocessing is a necessary phase that includes various steps aimed at reducing noise and preparing the data for various processing modalities. The following practices were applied to the acquired data:

\subsubsection{Radiometric and geometric corrections}
Each scanning session begins with calibration acquisitions. The first is a dark current acquisition via a closed shutter to correct the intrinsic sensor noise. The second is a white reference panel measurement for each camera. 
As the two sensors operate in different ranges of the electromagnetic spectrum — the FENIX sensor (VNIR-SWIR) measures purely reflected radiation, whereas the FX50 sensor (MWIR) captures a mixture of reflected and thermally emitted radiation —, each camera requires a reference panel of a different material suited to its respective spectral profile, as detailed in Table \ref{tab:parameters}.
These acquisition procedures allow for conversion of the raw digital numbers into absolute reflectance values, minimizing acquisition variables across different imaging sessions and environmental conditions. 
Radiometric and geometric corrections of the measurements were carried out using the LUMO Scanner software (LUMO Scanner, 2020) and the Hylite toolbox \cite{thiele2021multi}. To ensure the accurate co-registration of acquired RGB and hyperspectral images, calibration images containing six fiducial markers were used to determine camera offsets and compute affine transformations \cite{thiele2024maximising}.


\subsubsection{Background removal}
Hyperspectral images acquired on a moving conveyor belt contain foreground (the objects of interest) and background (conveyor belt or a table) pixels. Typically, the undesired background pixels need to be masked, as processing them adds computational cost and confuses the classifier. This is particularly the case when the spectral signature of the background material is similar to the objects of interest, i.e. dark-plastic conveyor belts. 

A foreground-background segmentation was thus applied, using the high spatial-resolution (166 micron per pixel) RGB images and segmentation workflow described in \cite{arbash2023masking}. This approach uses a 3-band image (RGB or false-colour image of HSI data cube) as input to Meta's Segment Anything Model \cite{kirillov2023segment}, which segments all detected objects in the image. Both a foreground semantic mask and an instance mask of all the image's objects are generated.
Semantic masks only distinguish foreground from background, while instance masks separate all individual instances. 
The same input is also processed using the large vision-language model, GroundingDINO \cite{liu2024grounding}, for zero-shot object detection of the group of objects in the image. The generated bounding box (GroundingDINO) and semantic mask and instance polygons (Segment Anything Model) are overlapped to derive object semantic and instance masks that separate plastic particles from background pixels and other false object detection near the image edges. These refined masks are then used to obtain a masked hyperspectral image containing the shredded objects only.

Semantic masks only distinguish foreground from background, with no inherent boundary separating individual instances.

\subsubsection{Normalization: Standard Normal Variate}

\begin{figure}[htbp]
     \centering
     \begin{subfigure}[b]{0.49\textwidth}
         \centering
         \includegraphics[width=\textwidth]{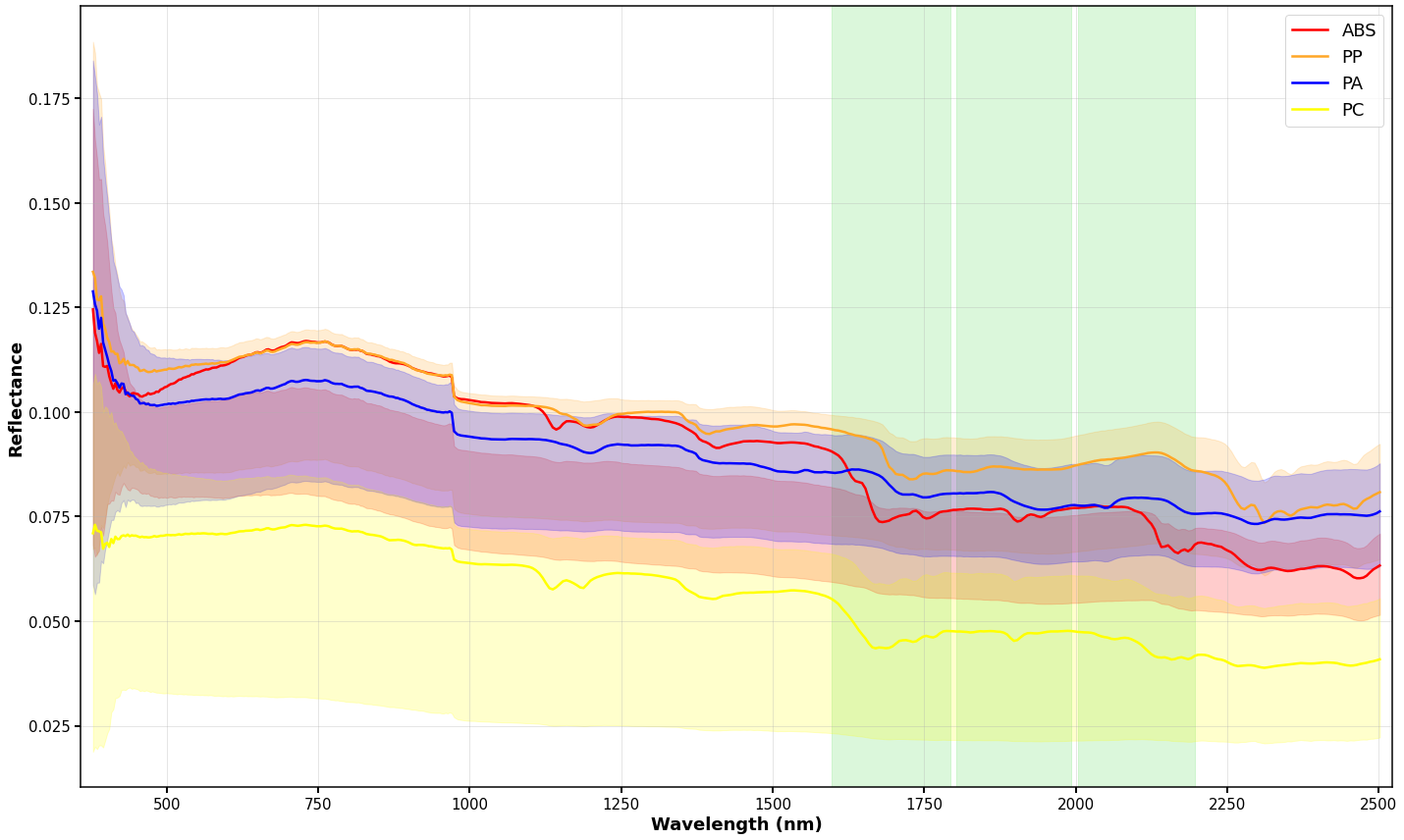}
         \caption{VNIR-SWIR: raw spectra}
         \label{HSI-SWIR Raw}
     \end{subfigure}
     \hfill
     \begin{subfigure}[b]{0.49\textwidth}
         \centering
         \includegraphics[width=\textwidth]{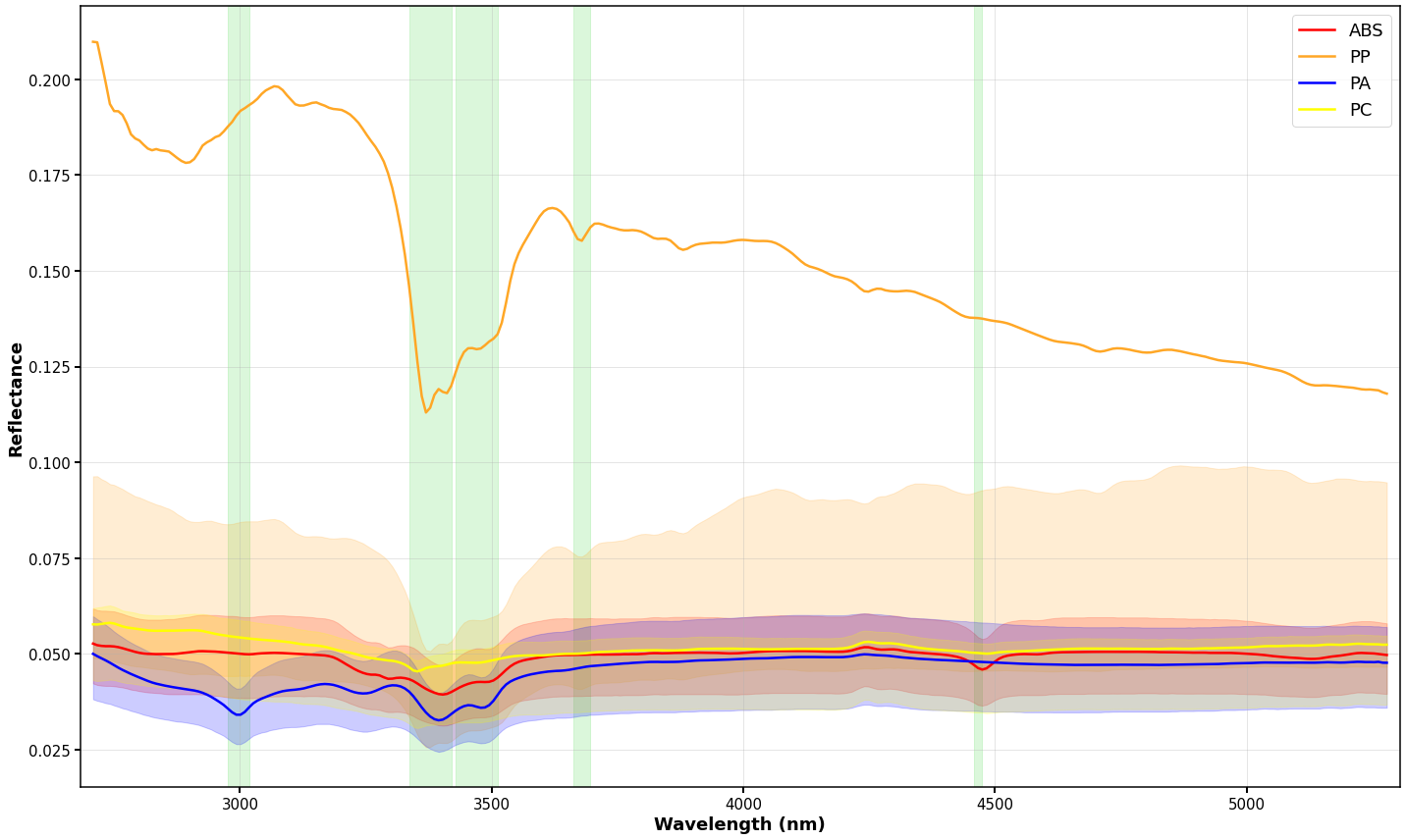}
         \caption{MWIR: raw spectra}
         \label{HSI-MWIR Raw}
     \end{subfigure}
     
     \vspace{1em} 
     
     \begin{subfigure}[b]{0.49\textwidth}
         \centering
         \includegraphics[width=\textwidth]{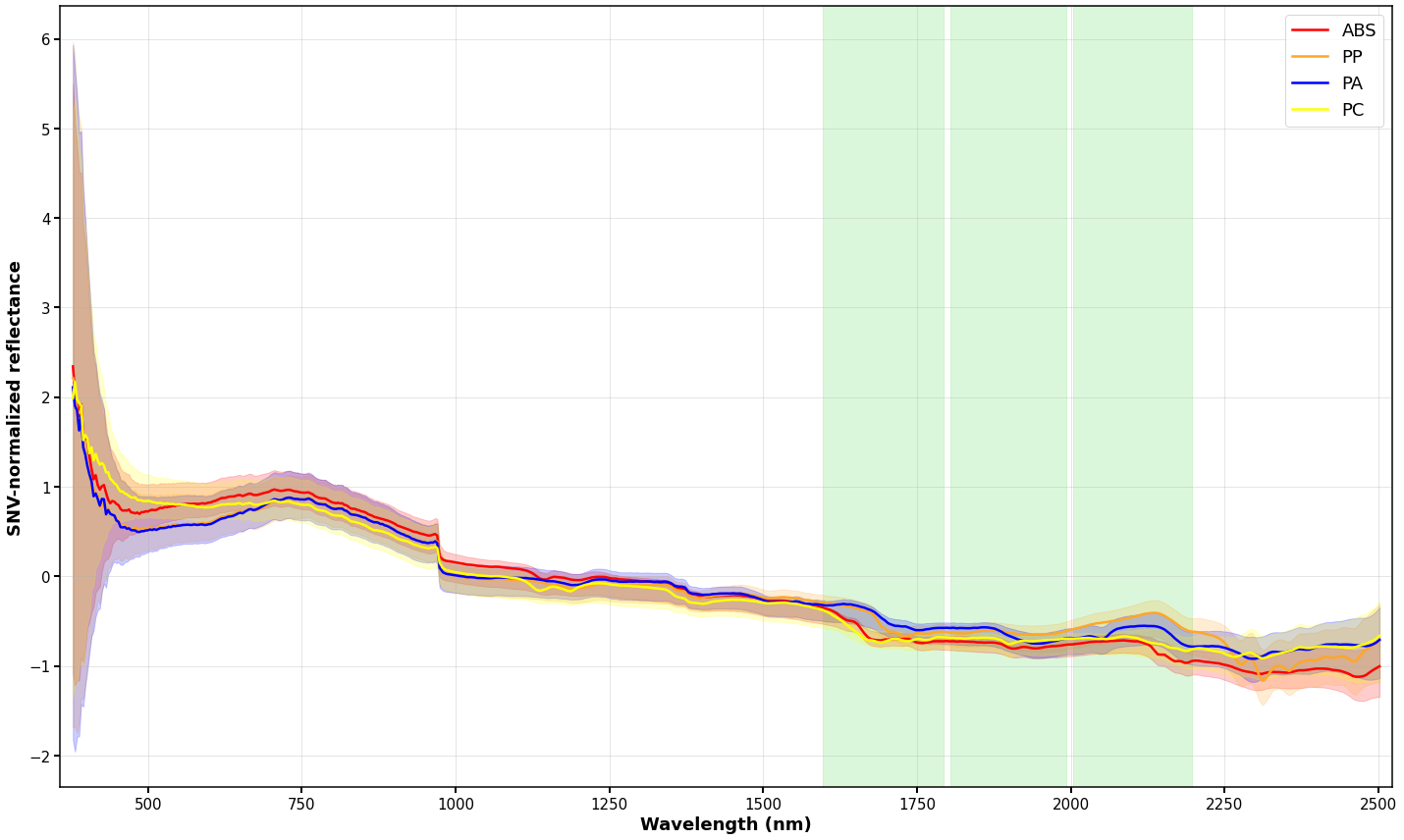}
         \caption{VNIR-SWIR: normalized spectra}
         \label{HSI-SWIR Norm}
     \end{subfigure}
     \hfill
     \begin{subfigure}[b]{0.49\textwidth}
         \centering
         \includegraphics[width=\textwidth]{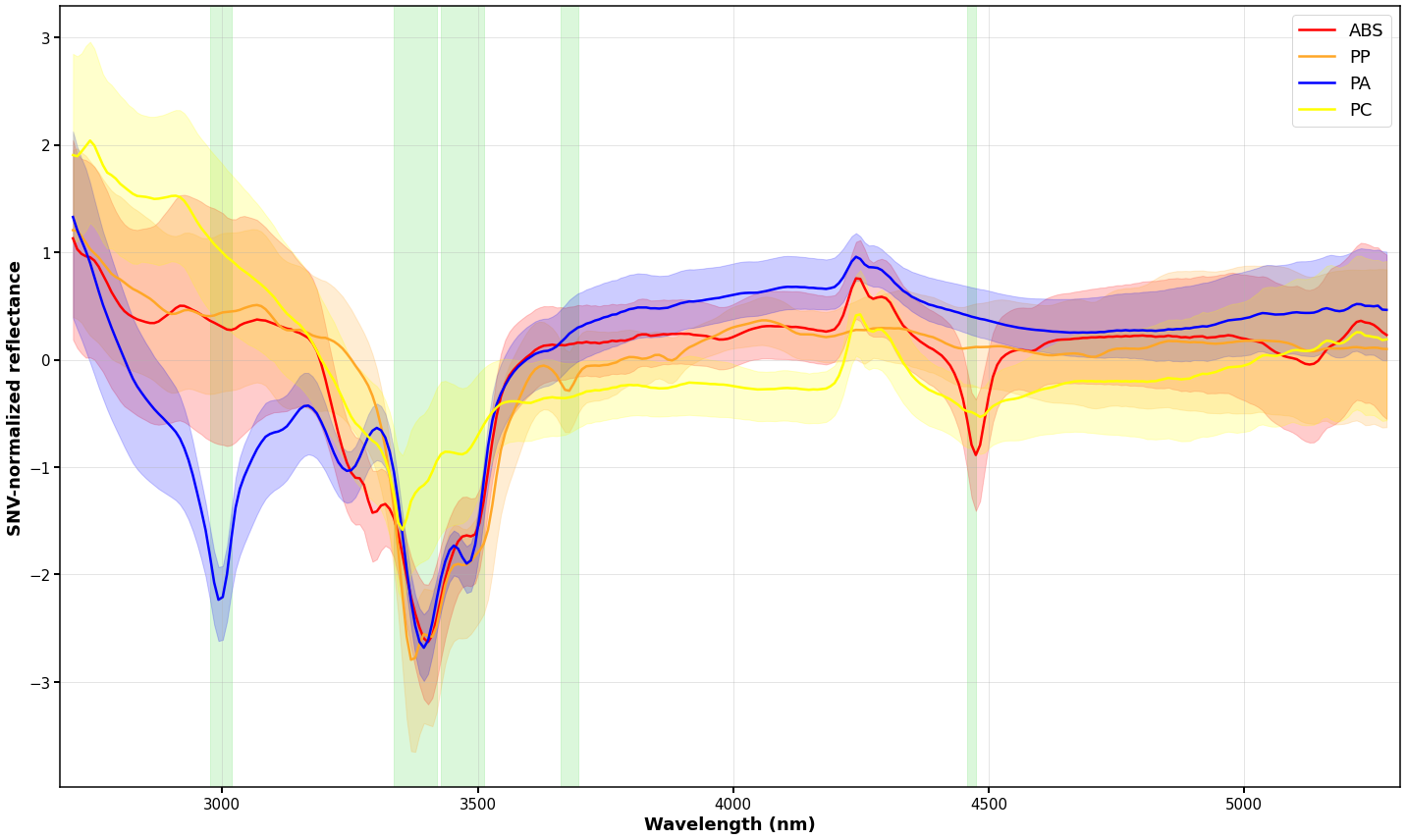}
         \caption{MWIR: normalized spectra}
         \label{HSI-MWIR Norm}
     \end{subfigure}
     
     \caption{The average spectra of the studied polymers including the 25\% and 75\% percentiles before (A and B) and after (C and D) SNV normalization.}
     \label{fig:spectra}
\end{figure}

The measured spectral signature is a function of the material's chemical composition and physicochemical state, including intrinsic chemical composition, surface roughness, surface geometry, illumination conditions, and sensor response. Since the proposed dataset consists predominantly of shredded carbon-black plastics together with particles of different colours and surface characteristics, considerable variations in absolute reflectance can occur within the same polymer class, as is the case for polymer PP in Figure \ref{fig:spectra} - \ref{HSI-MWIR Raw}. This variability is further aggregated by the different sensing principles of the two hyperspectral modalities: black plastics exhibit low, nearly featureless reflectance in the SWIR region, whereas in the MWIR region both reflected and thermally emitted radiation occur, resulting in larger reflectance variations and different absolute reflectance scales (due to sensor-specific calibration panels). These effects risk biasing classification toward noise-sensitive intensity thresholds rather than the robust, absorption-feature-based identification. 
Moreover, in our processing approach, we treated the spectra of the two sensors as a single data input for the classification methodologies by concatenating the two sensors' spectra at each pixel along the band axis. Therefore, we applied independent normalization of each sensor before the concatenation to subject both regions to a single shared mean and variance, ensuring that the scale and variance of one spectral region do not distort or dominate the features of the other. 
Accordingly, we applied standard normal variate (SNV) normalization to each spectrum, $x(\lambda)$, to compensate for multiplicative ($\alpha$) and additive ($\beta$) scatter effects on the true spectrum, $s(\lambda)$. SNV rescales the spectrum into a zero mean and a unit variance while preserving the relative absorption characteristics, thereby removing scatter-induced baseline and slope variations as seen in Figure\ref{fig:spectra} (\ref{HSI-SWIR Norm} and \ref{HSI-MWIR Norm}) :
$$
x(\lambda)=[x_{\lambda_1} ,x_{\lambda_2} ,…,x_{\lambda_b}], \qquad x(\lambda) = \alpha \cdot s(\lambda) + \beta
$$
The normalized spectrum is
\begin{equation}
x_i^{\mathrm{SNV}} = \frac{x_i - \overline{x}}{\sigma},
\qquad i = 1,2,\ldots,b,
\end{equation}
where $\overline{x}$ and $\sigma$ are the mean and standard deviation of that single spectrum across all spectral bands $b$.


\subsubsection{Band selection and concatenation}
The combined complete bands from the two spectral sensors present a challenge for the classification task, due to the curse of dimensionality and computational overhead. The high dimensionality presents especially significant overfitting challenges. The objective of hyperspectral image analysis is thus often to reduce data dimensionality while retaining sufficient information to identify, classify, and quantify the physicochemical properties of a sample \cite{burger2011data}. Consequently, following the normalization phase, a spectral subset was selected for input into the processing model, based on knowledge of the spectral regions relevant to polymer identification (green regions in Figure \ref{fig:spectra}). These subsets are based on the findings of De Lima Ribeiro et al. \cite{de2024multi} in the VNIR-SWIR range, whilst band selection for the MWIR range followed the expected fingerprint bands outlined in Table \ref{tab:parameters}. 
Finally, to derive a single (per-pixel) feature vector for the classification models, we concatenated the selected bands from the HSI-SWIR and HSI-MWIR datasets along the spectral (band) axis, combining the 107 bands of the HSI-SWIR (FENIX) cube with the 36 bands of the HSI-MWIR (FX50) cube for a total of 143 bands. This band-selection step reduced the models' input from 758 to 143 bands, improving speed and reducing overfitting by focusing attention on physically meaningful features.

\subsubsection{Train-Test Split}
Evaluating data-driven models, such as ML models, starts with training the model on a training set, followed by evaluation on a held-out test set. To ensure unbiased evaluation, we split the dataset at the scene level rather than the pixel level, with the test set representing challenging real-world sorting conditions. 
Scenes containing objects of size 5 mm to 11 mm were assigned to the test set, while scenes containing 5 mm to 30 mm objects were used for training. The result is five test scenes, four single-class polymer scenes — one scene for each class (ABS, PA, PC, PP), and one scene with mixed classes, with a total of 375,771 test vectors (pixels/patch), and eight training scenes with a total of 651,186 training vectors. 

\begin{figure}
    \centering
    \includegraphics[width=0.75\linewidth]{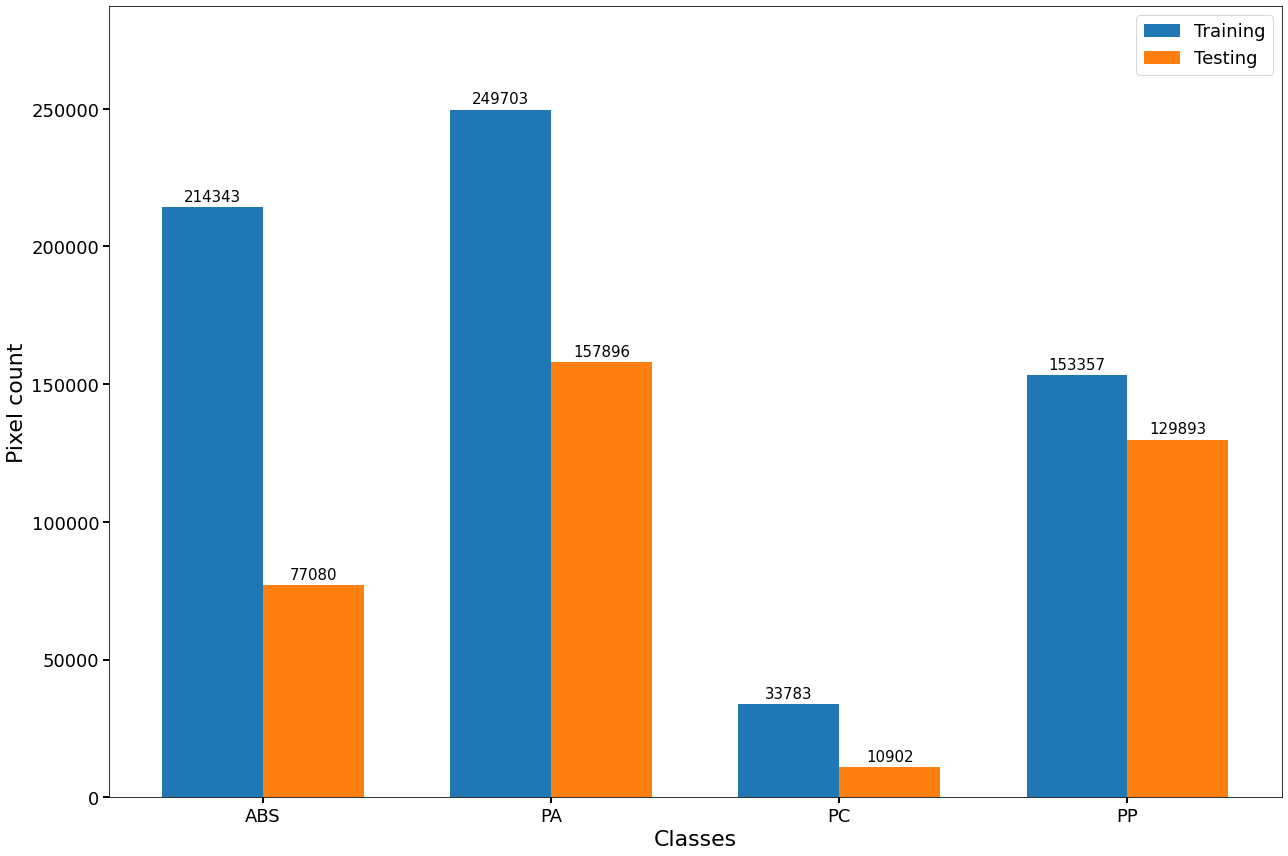}
    \caption{Train-test split class distribution.}
    \label{fig:stats}
\end{figure}

The train-test set sample distribution indicates an imbalance in the class distribution between the four classes. Although the least-appearing class (PC) has a total of 33783 training vectors (Figure \ref{fig:stats}), the imbalance compared to other classes such as ABS and PA can negatively bias the model training towards the most frequent class. To mitigate this issue, we applied a weighted cross-entropy loss function, where class weights are computed inversely proportional to their frequencies in the training set. This approach rebalances the contribution of each class by assigning greater penalties to misclassifications of underrepresented classes, thereby reducing the dominance of majority classes during training.

\subsection{Benchmarking experiments}
We compared several different spectral processing methodologies for automated classification of the four complex polymers presented in the provided dataset, including classical similarity-based approaches such as spectral angular mapper (SAM), statistical ML algorithms, and DL neural networks such as convolutional and transformer architectures. Each of these benchmarked methods is described in the sections below.

\subsubsection{Spectral Angular Mapper} 
SAM treats each spectrum as a unit vector in an $n$-dimensional space, where $n$ is the number of spectral bands (143 in our case). For each polymer class, a reference spectrum was computed as the mean of all training vectors belonging to that class. Given a test spectrum $\mathbf{t}$ and a reference spectrum $\mathbf{r}$, SAM computes the spectral angle $\alpha$ between the two vectors as:
\begin{equation}
    \alpha = \arccos \left( \frac{t \cdot r}{\|t\| \|r\|} \right)
\end{equation}
SAM is inherently insensitive to illumination differences because it measures only the angular separation between vectors. Each test spectrum was compared with every class-mean reference spectrum and assigned to the class that yielded the smallest spectral angle, i.e., the highest spectral similarity.

\subsubsection{Machine Learning}
\begin{itemize}
    \item Support vector machine (SVM) seeks the hyperplane that optimally separates the four plastic classes in the feature space. When classes are not linearly separable in the original 143-dimensional spectral space, nonlinear kernels implicitly map the spectra into a higher-dimensional feature space in which nonlinear boundaries can be represented. The decision function is defined by a subset of training samples, known as support vectors, whose kernel-based similarities capture spectral differences across the full spectral profile, including subtle variations in absorption features. Rather than relying on individual absorption bands, SVM considers the complete spectral signature, allowing differences in absorption depth, band shape, and shoulder position to contribute jointly to polymer discrimination.
    \item Random forest (RF) splits the training spectra using thresholds on the reflectance values of individual spectral bands, selecting at each node the band and threshold that most reduces class impurity. The algorithm's search for the most discriminative splits implicitly performs a form of feature selection, tending to select wavelengths that capture prominent spectral differences across the training data distribution. Averaging predictions across many trees, each trained on a bootstrap sample and a random subset of bands, reduces the variance associated with any single split, providing robustness against noisy or redundant spectral information. 
    
    \item K-nearest neighbors (KNN) classifies a test spectrum by directly measuring its distance, typically Euclidean, to every training spectrum in the $n$-dimensional band space, and assigning the class of the majority of $k$ nearest neighbors. Unlike SVM or RF, KNN builds no explicit model of the class boundary, it assumes that spectra of the same polymer class cluster locally in reflectance space because they share similar absorption-feature magnitudes and positions across the spectral signature. 
    \item Linear discriminant analysis (LDA) is a supervised dimensionality reduction technique that extracts the linear projection which maximizes the ratio between-class variance \cite{mclachlan2004discriminant}. First, LDA computes a small number of discriminant axes (linear combinations of all bands) along which the four polymer classes are maximally separated, classification is then performed in the reduced space. 
\end{itemize}

\subsubsection{DL}
\begin{itemize}
    \item 1D Convolutional Neural Networks (1D-CNN) treat the spectrum as one sequence instead of independent bands and apply learned convolutional filters that slide across the spectral signature in sliding windows. Each filter is optimized during training to respond to specific local spectral shapes rather than to the reflectance value at any single band in isolation. Successive convolutional layers combine these local detections into increasingly abstract representations. This convolutional structure gives 1D-CNN an explicit inductive bias toward local spectral continuity that classical ML models lack, as most treat each band as an independent input feature without an inherent assumption of relationship between neighboring wavelengths. The architecture of the 1D-CNN was adopted from \cite{xia2021convolutional}.
    \begin{equation}
H^{(l)} = \phi\!\left(W^{(l)} * H^{(l-1)} + b^{(l)} \right),
\qquad
H^{(0)}=x,
\end{equation}
$H^{(l)}$ is the feature map produced by the ( $l_{th}$) convolutional layer. $\phi(\cdot)$ is the ReLU nonlinear activation function applied to the convolution operation ($*$)  between the learnable 1D convolutional kernels $W^{(l)}$ and the feature map of the previous layer (${l-1}_{th}$), added to the bias term $b^{(l)}$.  

    \item Vision Transformer (ViT) \cite{dosovitskiy2021an} extends the sequence-based learning of 1D-CNNs by replacing local convolutional filters with a self-attention mechanism \cite{vaswani2017attention} that relates every spectral position to every other, regardless of distance along the band axis. The spectrum (or a spectral patch) is first divided into a sequence of tokens, each linearly embedded and combined with a positional encoding to preserve the wavelength order; self-attention then computes learned weights over all pairs of tokens. This is a fundamentally different inductive bias from the 1D-CNN's local receptive field, which assumes that the informative structure is local and shape-based. Transformers' self-attention assumes that class-discriminative information may depend on joint patterns across multiple, non-adjacent absorption features simultaneously.
    \item Spectral Former (SF) \cite{hong2021spectralformer} adapts the transformer architecture specifically to hyperspectral data by grouping neighboring bands into single token embeddings prior to attention, in a tokenization step called group spectral embedding. In the pixel-wise variant, this grouping and attention mechanism operates over a single spectrum's band sequence, capturing global-range dependencies between grouped absorption regions. In patch-wise SF, spatial neighborhoods of pixels are additionally incorporated, a patch of size  $9\times9 $ is applied, allowing the model to leverage spatial-spectral context, i.e., the neighboring pixels on the same shredded object are likely to share the same polymer class. 
    \begin{equation}
H^{(l)}
=
\mathrm{FFN}
\left(
\mathrm{Softmax}
\left(
\frac{QK^{\top}}{\sqrt{d}}
\right)
V
\right),
\qquad
H^{(0)}=E+P,
\end{equation}
After the input spectrum $x$ is tokenized and embedded into an another learnable vector $E$, positional embedding $P$ is added element-wise to the token embeddings. At each Transformer layer that follows, the input representation is projected into the queries $Q$, keys $K$ and values $V$ learned matrices where the affinities between every query and every key is computed through the scaled dot product $\frac{QK^{\top}}{\sqrt{d}}$, producing the similarity matrix that quantifies how strongly each spectral band is linked to every other. The $Softmax$ function is then applied to convert the similarities into normalized weights, allowing the model to emphasize the most informative spectral relationships. These attention weights are multiplied by the value matrix $V$, producing a weighted combination of information from all spectral bands. Finally, the resulting representation is refined by a feed-forward network (FFN). Through repeated application of this workflow, successive layers learn increasingly abstract representations that capture both local and long-range spectral dependencies.
\end{itemize}

For the ML models, hyperparameters were optimized by evaluating 25 randomly sampled configurations for each classifier (random search). The full parameter ranges and grids are provided in the released code. 
The DL models were trained using the Adam optimizer with a learning rate of $1\times10^{-7}$,  a mini-batch size of 512, a maximum of 600 training epochs, and a weighted categorical cross-entropy loss to account for class imbalance. For SpectralFormer, the architectural hyperparameters were adopted from the original implementation, similar to the group spectral embedding size of 7. To account for training stochasticity, each DL model was trained three times using different random initialization seeds, and the reported performance corresponds to the average across these runs.  The averaged independent training runs exhibited low variability, with a maximum deviation of less than $0.03$ from the mean performance, indicating that the reported results are robust to random weight initialization. For all models, the main results (Figure~\ref{fig:bar}) report the average F1-score of the five best-performing hyperparameter configurations ranked by the average Cohen's $\kappa$. 

\subsection{Post-processing}

Following the prediction of the test-set vectors, mask construction is applied to build the prediction map of the hyperspectral scene from individual vectors. The result is a pixel-wise classification mask, where each pixel is assigned the class predicted by the classifier for that vector. Following this, a majority voting process is applied to the pixel-wise classification masks within the objects' polygons in the semantic and instance segmentation masks generated during the preprocessing phase. This projection, as shown later, suppresses noisy and spurious pixel-level predictions and aggregates them into object-level classifications. Depending on the mask used, the projection can operate at either the semantic or instance level. 
Semantic projection is computationally efficient, as majority voting is performed over a small number of foreground/background regions, making it well suited for real-time applications. In contrast, instance projection performs majority voting independently within each segmented object, preserving instance awareness and enabling object-level classification at the expense of higher computational cost. 

\section{Results}
\label{sec:experiements}

\begin{figure}
    \centering
    \includegraphics[width=\linewidth]{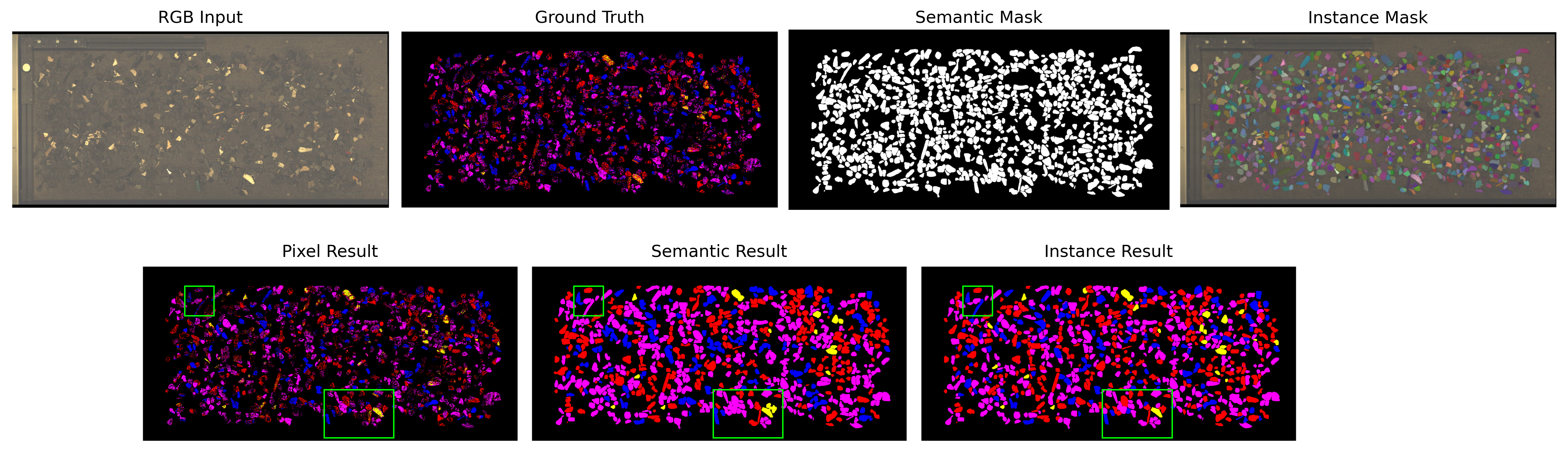}
    \caption{Overview of the scene processing pipeline. The RGB image is segmented using GroundingDINO and SAM to generate semantic and instance masks. Classifiers are trained and validated using the pixel-wise ground truth across four classes: Styrene (\colorbar{Class_1!}), PA (\colorbar{Class_2!}), PC (\colorbar{Class_3!}), and PP (\colorbar{Class_4!}). The resulting pixel-wise predictions are then aggregated via majority voting within each mask to produce semantic- and instance-level classifications. Green boxes highlight cases where instance segmentation resolves overlapping or touching objects that semantic segmentation merges into a single prediction.}
    \label{fig:pred}
\end{figure}

Confusion matrices were computed for the predicted masks at each classification stage, and model performance was quantified primarily against the pixel-wise ground truth using the F1-score and Cohen's Kappa. F1-score was selected as the primary metric because it balances precision and recall, jointly penalizing both missed detections (false negatives) and incorrect predictions (false positives). Cohen's Kappa was reported to quantify the degree of agreement between predicted and ground-truth labels beyond random classification. Kappa ranges from  $-1$ to $1$, where $1$ indicates perfect agreement, $0$ indicates complete randomness, and negative values indicate disagreement.

\begin{figure}[htbp]
     \centering
     \begin{subfigure}[b]{0.49\textwidth}
         \centering
         \includegraphics[width=\textwidth]{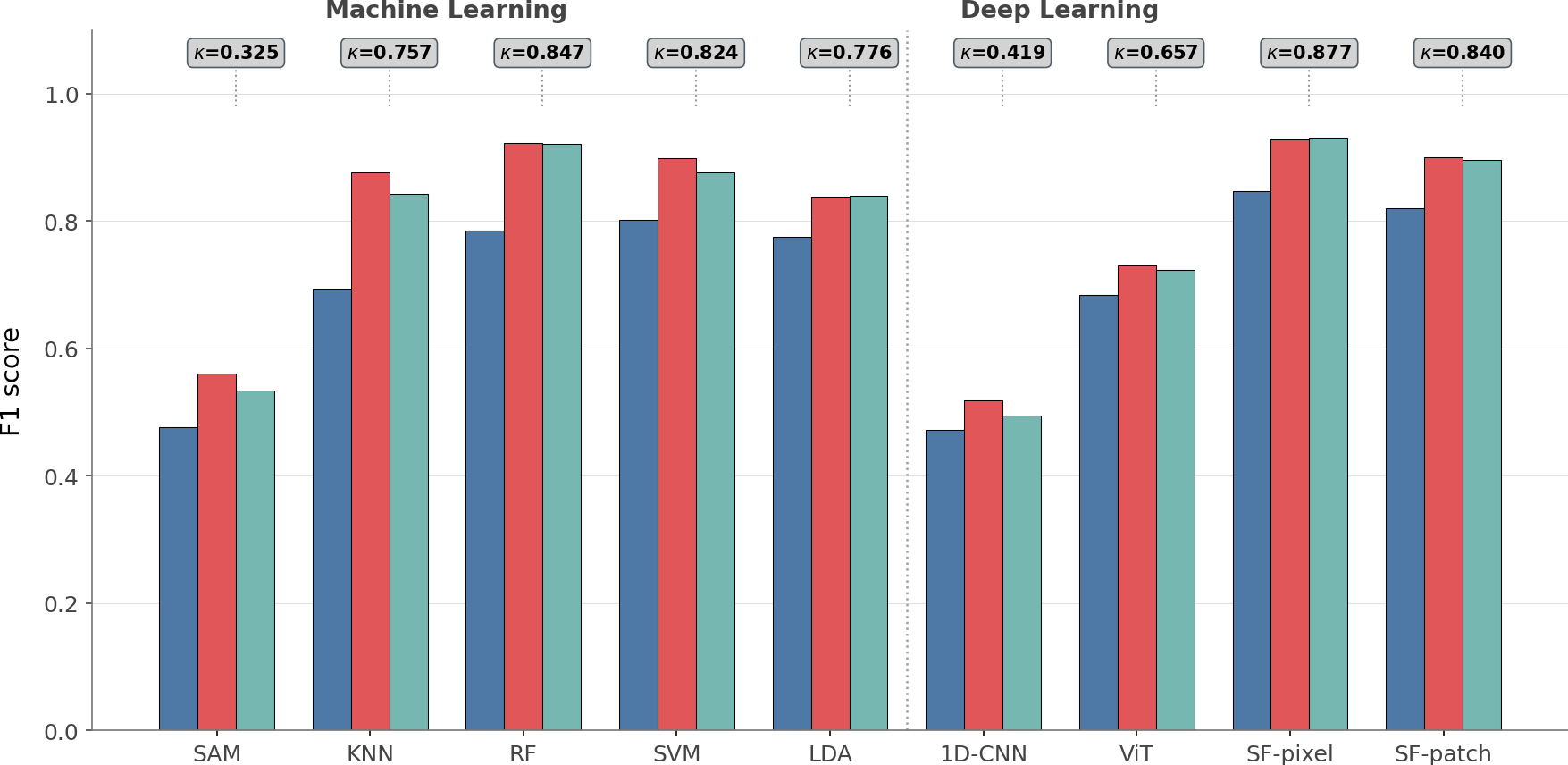}
         \caption{Styrene}
         \label{Styrene}
     \end{subfigure}
     \hfill
     \begin{subfigure}[b]{0.49\textwidth}
         \centering
         \includegraphics[width=\textwidth]{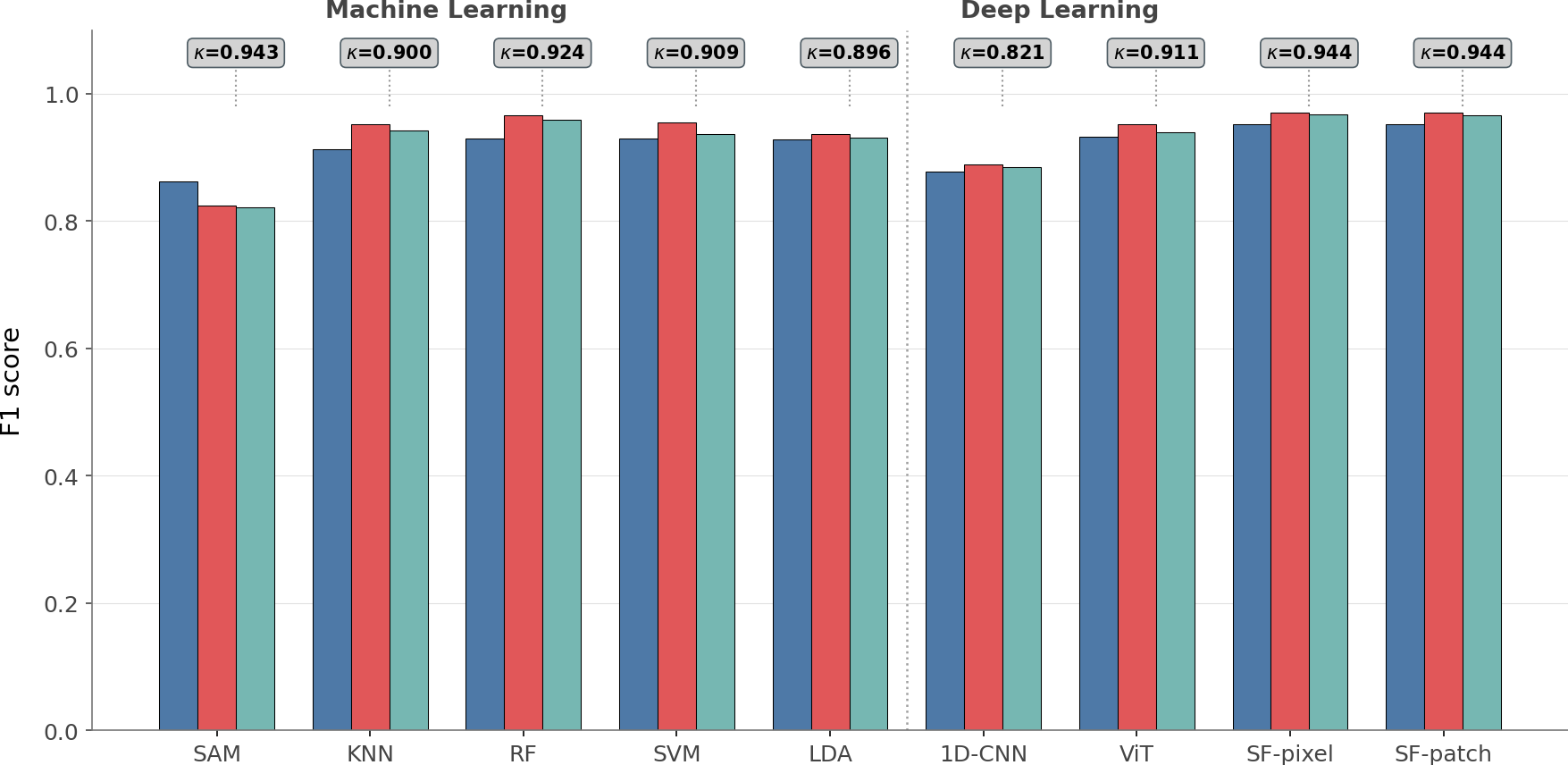}
         \caption{PA}
         \label{PA}
     \end{subfigure}
     
     \vspace{1em} 
     
     \begin{subfigure}[b]{0.49\textwidth}
         \centering
         \includegraphics[width=\textwidth]{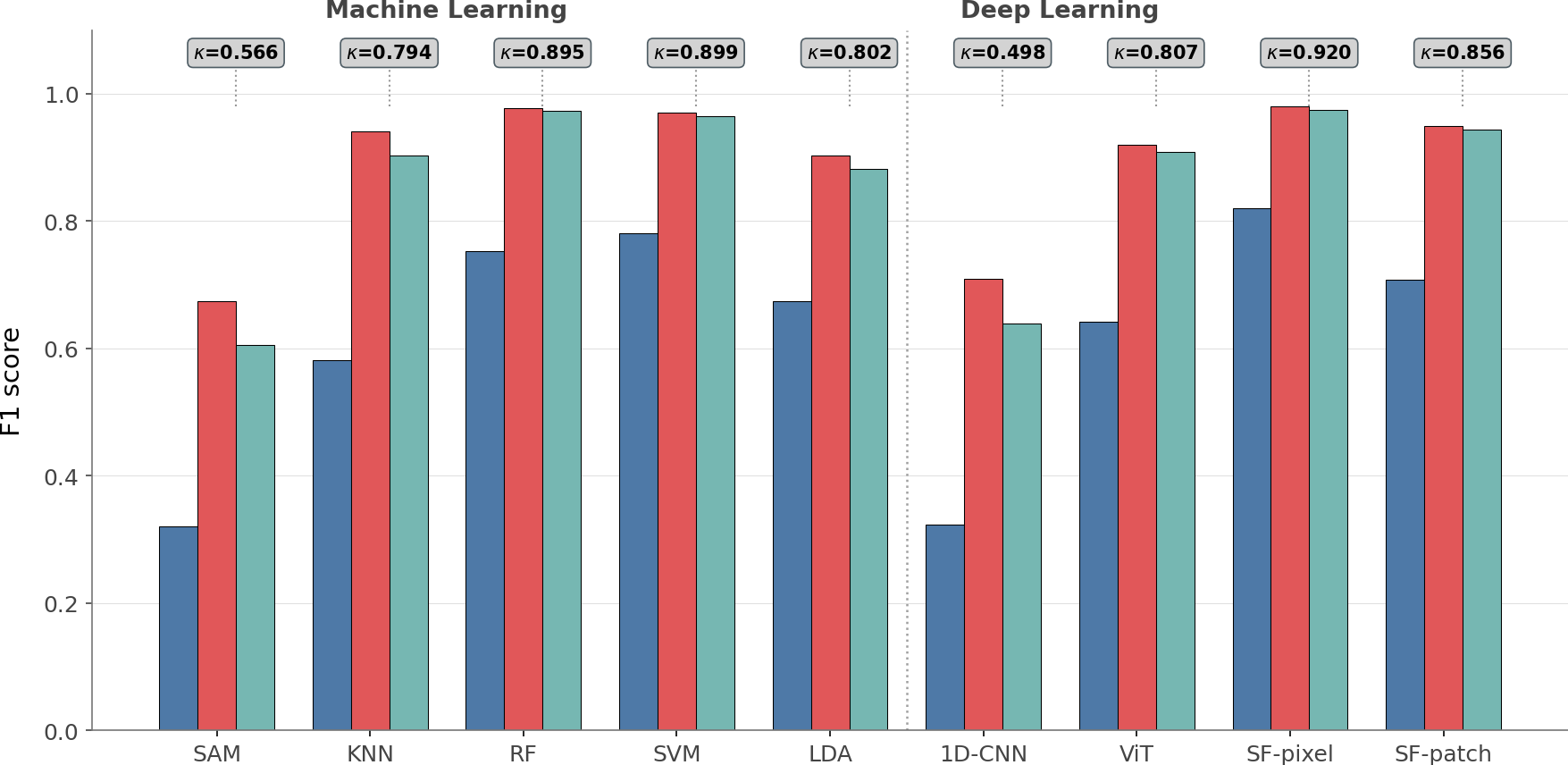}
         \caption{PC}
         \label{PC}
     \end{subfigure}
     \hfill
     \begin{subfigure}[b]{0.49\textwidth}
         \centering
         \includegraphics[width=\textwidth]{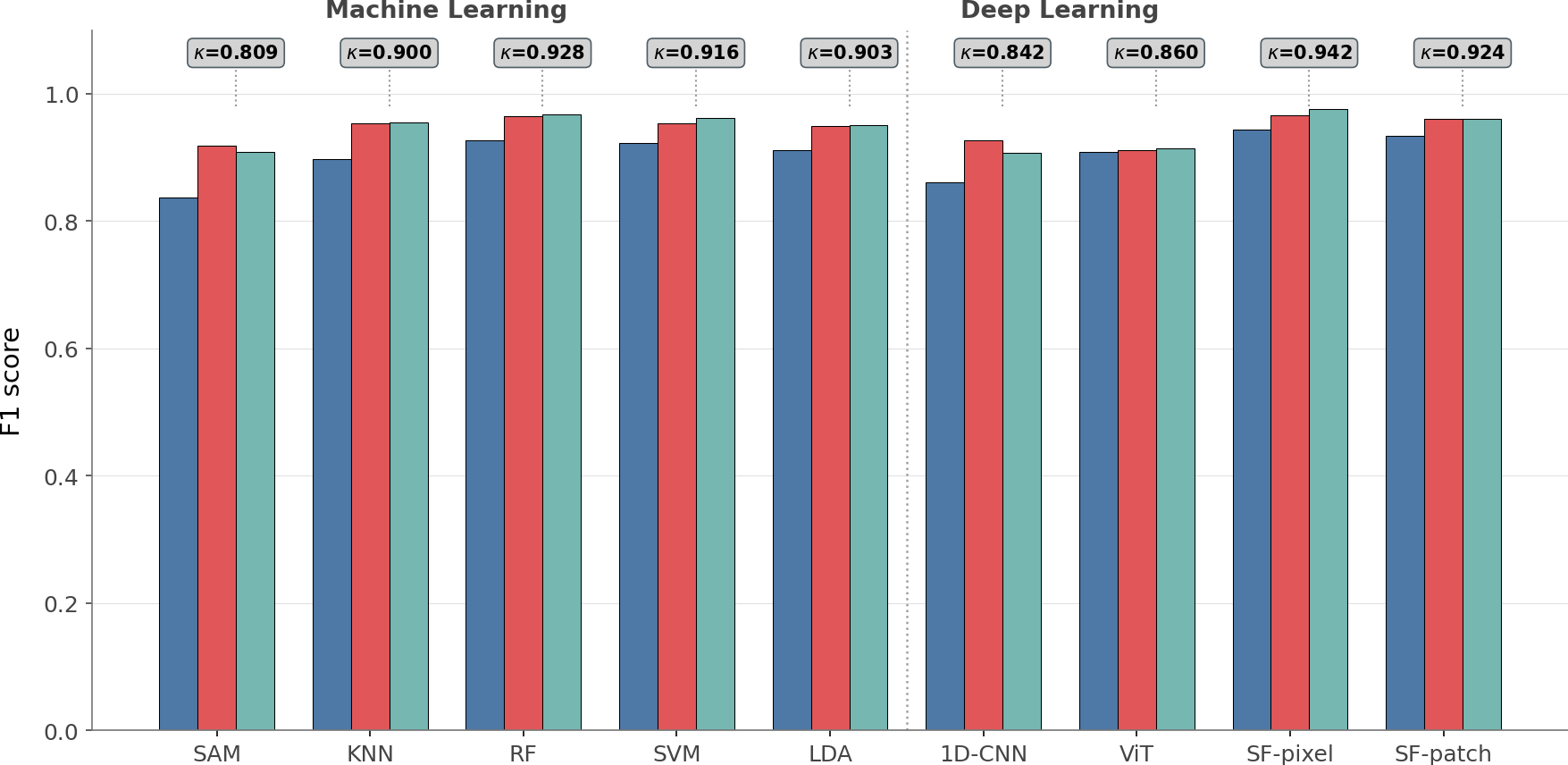}
         \caption{PP}
         \label{PP}
     \end{subfigure}
     
     \caption{The F1-score and Cohen's $Kappa$ for each of the tested models across the four classes and scenes in the test set. The blue (\colorbar{blue!}), red (\colorbar{red!}), and green (\colorbar{green!}) bars correspond to pixel-wise classification, semantic object segmentation, and instance-level segmentation, respectively.}
     \label{fig:bar}
\end{figure}

The per-class F1 scores in Figure \ref{fig:bar} revealed heterogeneity in classification performance across the four classes. PA in Figure \ref{PA} and PP in Figure \ref{PP} were consistently the best-recognized classes, with F1 scores exceeding 0.83 even at the pixel-wise classification and reaching $0.91$– $0.98$ across most classifiers after semantic or instance aggregation; furthermore, the performance for these two classes (PA and PP) was insensitive to the majority voting type. PC in Figure \ref{PC} exhibited the opposite behavior: pixel-wise F1-score was both low and highly variable across classifiers (0.32–0.78), whereas semantic-object aggregation voting produced the largest per-class gain, indicating that PC misclassifications were dominated by pixel-level noise which was resolved by spatial aggregation. Styrene in Figure \ref{Styrene} remained the most challenging class at every stage, with pixel-wise F1 as low as $0.47$–$0.52$ for the weakest classifiers, but with recovered performance after aggregation, ceiling at $0.90$–$0.93$ for the best (SpectralFormer-pixel, Random Forest).

\subsection{Pixel-wise Classification}

Our dataset is challenging for classification, featuring 143 selected bands, weak absorption in SWIR for carbon-black plastics (Figure \ref{HSI-SWIR Raw}), prominent absorption features in MWIR (Figure \ref{HSI-MWIR Norm}, Table \ref{tab:parameters}), and reflectance variations caused by particle orientation, surface roughness, illumination, and emissivity. 
Pixel-wise classification results offer valuable insights into individual classifier performance and the compatibility of data distributions between training and test sets.

Among the models, SF-pixel achieves the highest F1-score across all classes. Its patch-wise counterpart and ViT, however, fall behind despite incorporating spatial context through patched input. More broadly, SF-pixel, LDA, RF, and SVM were the top performers across all classes and evaluation stages, while the SAM and 1D-CNN were consistently the weakest. RF and SVM, in particular, achieve over 85\% F1-score performance.

Furthermore, the model size and inference speed differ substantially across architectures; the high-performant RF model requires $1.3$ GB of storage compared with less than $1$ MB for SF-pixel. On a high-performance computing platform with an NVIDIA A100 GPU, all models completed inference in under $1$ ms per pixel. 


Given the dataset characteristics described above, discriminative information arises both locally, within individual absorption features, and globally, across distant spectral regions, favoring models that jointly capture local and long-range dependencies. 
The superior performance of SF-pixel indicates the advantage of the Transformer architecture over the other models. 

While spatial features enrich the representation in principle, spectral mixing at object edges (where neighboring pixels from adjacent objects or the background contribute to a single patch) likely introduces confusing information, explaining the relative underperformance of the patch-wise SF and ViT. 


The near-flat, low-absorbance spectra of carbon-black plastics (Figure \ref{HSI-SWIR Norm}) yield spectral angles that are highly similar across classes — when compared to each class's reference spectrum — reducing the discriminative ability of SAM.

CNNs' convolutional filters have a fixed, local receptive field and directly capture short-range dependencies, but require many stacked layers to progressively expand their receptive field toward longer-range relationships. Transformers' self-attention mechanism, on the contrary, directly models long-range spectral dependencies and global prominent features from the first layer.


ML models RF and SVM match the top-performing DL models. 
RF ignores irrelevant bands, learns nonlinear thresholds, and is robust to noisy spectra as it averages hundreds of trees.
SVM is capable of constructing maximum-margin decision boundaries. 


These results indicate that, when combined with appropriate preprocessing, multi-modal spectroscopy imaging enables successful classification of bulk carbon-black polymers, particularly when paired with advanced ML and DL models capable of learning complex decision boundaries and long-range spectral dependencies.
However, since HSI acquisition and data transfer/storage constitute the dominant bottleneck in the current acquisition-processing workflow, the processing-time differences reported above are not yet the limiting factor for deployment. 

More broadly, the consistently strong performance of both ML and DL models across SWIR, MWIR, and RGB modalities is consistent with a close similarity in data distribution between the training and test scenes, i.e., the model in the training set got introduced to the type of objects in the test set. 
Because the spectral signature encodes both the materials' intrinsic chemical composition and other surface characteristics (e.g., roughness), applying these trained models to a different dataset may not yield comparable performance. Retraining, or fine-tuning, on a representative subset of new data is therefore essential to maintain reliable classification performance in deployment settings beyond those examined in this study.


\subsection{Object-wise Classification}

\subsubsection{Semantic Object Classification}

The difference observed between the semantic object classification (red bars) and pixel classification (blue bars) in Figure \ref{fig:bar} indicates a substantial improvement in classification performance when pixel-wise predictions are aggregated into object-level predictions. The background masking, performed in the preprocessing stage using the high-spatial-resolution RGB modality, isolates foreground objects in a single binary mask. Majority voting thereafter assigns all pixels within each connected foreground region to the class label predicted most frequently among its constituent pixels. 
This approach compensates for the lack of spatial context in HSI pixel-wise classifiers by incorporating object-level spatial information at the post-processing stage, consequently mitigating the false predictions of the pixels near object boundaries and their light scattering effects that weaken the classification accuracy of pixel-wise classifiers.


Semantic masks only distinguish foreground from background, with no inherent boundary separating individual instances. Consequently, when multiple objects overlap or touch, their pixels merge into a single connected foreground region as that of one object. Applying majority voting to this region therefore assigns the same label to what are, in fact, distinct objects. This is visible in Figure \ref{fig:pred} within the green boxes, where overlapping or touching objects are merged into a single foreground region. We mitigate this effect by applying majority voting on the instance masks, to obtain instance level classification.


\subsubsection{Instance Object Classification}

The green boxes in Figure \ref{fig:pred} highlight the advantage of instance segmentation over semantic segmentation in the case of overlapping objects; unlike semantic segmentation, which produces binary foreground masks, a single one for touching objects, instance segmentation assigns a separate mask for every object or sub-object. As a result, it allows each object to retain its own prediction, proving the suitability of instance segmentation for identification tasks of overlapping objects.

However, from Figure \ref{fig:bar}, instance-level performance (green bars) is seen to fall slightly below semantic-level performance (red bars), while still remaining higher than pixel-wise performance (blue bars).
The reason for this gap stems from the fact that the zero-shot segment-anything model over-segments a single object into multiple instances, causing independent voting. This over-segmentation tendency is a known limitation of zero-shot detectors and can be mitigated through fine-tuning of image foundational models on customized datasets related to test samples. Fine-tuning the segmentation stage remains a natural direction for future work, particularly for pipelines that rely directly on predicted instance masks for downstream evaluation.


Finally, we claim a more trustable polymer classification, one that stems directly from the two-step design at the core of our approach: one step handling spatial features, another step handling spectral features, each carried out by advanced models pulled from computer vision and HSI remote sensing domains. Particle classification itself relies purely on spectral information from HSI-SWIR and HSI-MWIR, without a possibility of the decision leaning on spatial shortcuts that carry no intrinsic material property.
Scattering and mixing effects at particle boundaries cause noisy edge behaviour. This is resolved in the second step, where spatial context is aggregated through majority voting, optimizing particle-level predictions.

\section{Discussion}
\label{sec:Discussion}

The multi-step classification sets our approach apart from predecessor plastic classification work. Where prior approaches largely stayed within a single classical modality and hand drawn region of interest, we deploy advanced DL architectures from hyperspectral earth observation into the close-range plastic identification domain via spectral processing, while pairing them with SOTA RGB segmentation foundation models for spatial aggregation, refining segmentation accuracy down to the scale of small, shredded particles. Underpinning both steps is in-house domain knowledge of plastic spectral behavior, built into a multi-scene dataset of tiny shredded pieces (longest axis smaller than $3$cm), ground truth extracted from intrinsic material features through range selection, and DL pipelines adapted specifically for this classification task. The fusion of two advanced modalities, each specialized in different feature extraction, moves black polymer identification a decisive step closer to reliable, deployable automated sorting.

The integration of multi-modal HSI with rapid DL methods represents the driving force towards a transformative approach for the accurate and high-throughput characterization of industrial dark polymer wastes. According to the literature studies, conventional VNIR and SWIR hyperspectral systems often fail to discriminate black or dark-colored plastics due to the strong absorption of this spectral range by carbon black pigments, which suppresses diagnostically useful spectral features. MWIR imaging has been shown to retain discriminative absorption features for carbon-black materials where VNIR and SWIR do not. Combining MWIR with SWIR within a multi-modal hyperspectral platform captures complementary information across a broader spectral range than either modality alone, mitigating the spectral "blind spots" of a single-sensor system.

Rapid DL architectures, including transformer-based models and lightweight networks can exploit this high-dimensional information to learn complex, non-linear mappings between spectra, textures, and material classes. Such models outperform rule-based methods such as SAM (\ref{fig:bar}) in both accuracy and generalizability,  
especially under real-world conditions involving soiling, surface degradation, and mixed polymer shapes. Embedding multi-modal fusion directly into the processing algorithm can dynamically weight modalities according to their information content for a given sample, thus, enhancing resilience to sensor noise and occlusion. This capability is critical for dark polymers, where subtle spectral differences in MWIR spectra may be the only reliable cues for distinguishing, for instance, PP from black ABS or carbon-black-filled PE. Moreover, rapid inference (under $1$ ms per pixel) enabled by prior band selection and hardware acceleration allows these systems to operate at industrial conveyor speeds, supporting real-time sorting decisions without compromising throughput. The synergy between multi-modal sensing and fast DL thus closes the gap between laboratory-grade spectroscopic fidelity and the speed requirements of recycling facilities. 
Nonetheless, spectral variance among black polymers arises from numerous sources beyond polymer type itself, including differences in carbon black loading, additive and filler composition, weathering or surface degradation, and surface geometry, each of which can shift or dampen absorption features of the polymer independently. While extensive, our dataset was collected from a defined set of scenes and material sources, and does not capture the full extent of black polymer variability encountered across various waste streams or processing facilities. Standard data augmentation techniques (e.g., spectral noise injection, scaling) can partially simulate acquisition variability but cannot substitute for the real variation in chemical composition or physical state. This requires genuine sample diversity to be adequately represented in training data.

Importantly, the data-driven nature of DL allows continuous improvement as new waste streams, additives, and contamination patterns emerge, provided that representative, well-annotated multi-modal datasets are maintained. Together, multi-modal HSI and rapid DL form a scalable, adaptable, and high-performance analytical pipeline tailored to the specific challenges of dark polymer wastes. By leveraging the discriminative power of fused spectral–spatial information, this approach substantially increases sorting purity, recovery, and economic viability of polymer recycling. Ultimately, it provides a scientifically rigorous and industrially deployable foundation for transforming currently underutilized dark plastic fractions into high-quality secondary resources.

\section{Conclusion}
\label{sec:conclusion}

In this paper we present a hyperspectral dataset and machine learning approach that can support refined plastic sorting in recycling facilities, with a specific focus on shredded black plastics of end-of-life vehicles. We acquired and co-registered imaging data of ABS, PA, PC, and PP in black using different modalities, including RGB, shortwave- and midwave-infrared hyperspectral sensors, to ensure coverage of a wide range of physicochemical features. We then segmented the acquired scenes using zero-shot large-language vision models for isolating foreground objects from undesired background. This allowed the evaluation of various spectral analysis algorithms, including classical and machine learning algorithms such as spectral angular mapping, K-nearest neighbor, random forest, and support vector machine, in addition to different neural network architectures like CNNs and SpectralFormer; the latter proved its classification superiority over the other chemometric methods. We applied majority voting on the pixel-level classification results to obtain object-level identification, then used the instance segmentation masks to overcome the overlapping object scenarios and achieved over $94\%$ and $95\%$ average F1-score and Cohen's Kappa, respectively. Our findings demonstrate the potential of extended range hyperspectral data, especially in the MWIR, for fast identification (under 1 ms per pixel) of black polymers. However, we also caution that substantial training data covering all possible spectral variance of black polymers is needed before confident predictions can be made for unseen samples. 
Our dataset, models and associated results are thus provided in a FAIR and open access way, to further support advanced recycling and associated reduction in plastic pollution and waste.




\section{Availability of data}
The dataset is available on: 

\url{https://zenodo.org/records/22142947}

\section{Code availability}
The code for this work was performed using Python and the PyTorch framework and is available at:

\url{https://github.com/hifexplo/MWIR-4-Plastic}

The code for the masking method with zero-shot segmentation is available at: 

\url{https://github.com/hifexplo/Masking}. 



\small
\section{References}
\bibliographystyle{unsrt}
\bibliography{reference}

\end{document}